\documentclass[10pt, conference, letterpaper]{IEEEtran}
\usepackage{verbatim}
\usepackage{cite}
\usepackage{times}
\usepackage{amsmath}
\usepackage{graphicx}
\usepackage{graphics}
\usepackage{subfigure}
\usepackage{epsfig}
\usepackage{amsmath}
\usepackage{latexsym}
\usepackage{amsfonts}
\usepackage{amssymb}
\usepackage{paralist}
\usepackage{xspace}
\usepackage{mathrsfs}
\usepackage{amssymb}
\usepackage{color}
\usepackage[noend]{algorithmic}
\usepackage[ruled]{algorithm}

\DeclareMathOperator*{\argmin}{arg\,min}
\def\ie{\textit{i.e.}\xspace}

\def\eg{\textit{e.g.}\xspace}

\newtheorem{theorem}{Theorem}

\newtheorem{lemma}{Lemma}

\begin{document}

\title{Networked Stochastic Multi-Armed Bandits with Combinatorial Strategies}
\author{
\IEEEauthorblockN{Shaojie Tang}
\IEEEauthorblockA{University of Texas at Dallas} \IEEEauthorblockN{Yaqin Zhou}
\IEEEauthorblockA{Singapore University of Technology and Design}}
\maketitle

\begin{abstract}
In this paper, we investigate a largely extended version of classical MAB problem, called \emph{networked combinatorial bandit problems}. In particular, we consider the setting of a decision maker over a networked bandits as follows: each time a combinatorial strategy, e.g., a group of arms, is
chosen, and the decision maker receives a reward
resulting from her strategy and also receives a \emph{side bonus}
resulting from that strategy for each arm's neighbor. This is motivated by many real  applications such as on-line social networks where friends can provide their feedback on shared content, therefore if we promote a product to a user, we can also collect feedback from her friends on that product. To this end, we consider two types of side bonus in this study: \emph{side observation} and \emph{side reward}.  Upon the number of arms pulled at each time slot, we study two cases: \emph{single-play}  and \emph{combinatorial-play}. Consequently, this leaves us four scenarios to investigate in the presence of side bonus: Single-play with Side Observation, Combinatorial-play with Side Observation, Single-play with Side Reward, and Combinatorial-play with Side Reward.
For each case, we present and analyze a series of \emph{zero regret} polices where the expect of regret over time approaches zero as time goes to infinity. Extensive simulations validate the effectiveness of our results. 
\end{abstract}

\section{Introduction}
\label{sec-intro}
A multi-armed bandits problem (MAB) problem is a basic sequential decision making problem defined by a set of strategies.
At each decision epoch, a decision maker selects a strategy that involves a combination of random bandits or variables, and then obtains an observable reward.
The decision maker learns to maximize the total reward obtained in a sequence of decisions through history observation.
MAB problems naturally capture  the fundamental tradeoff between exploration and exploitation in sequential experiments. That is, the decision maker must exploit strategies that did well in the past on one hand, and explore strategies that might have higher gain on the other hand.
 MAB problems now play an important role in online computation under unknown environment, such as pricing and bidding in electronic commerce \cite{babaioff2012dynamic,babaioff2010truthful}, Ad placement on web pages \cite{chervonenkis2013optimization}, source routing in dynamic networks \cite{polychronopoulos1996stochastic}, and opportunistic channel accessing in cognitive radio networks \cite{li2012almost,zhao2008myopic}. 
In this paper, we investigate a largely extended version of classical MAB problem, called \emph{networked combinatorial bandit problems}. In particular, we consider the setting of a decision maker over a networked bandits as follows: each time a combinatorial strategy, e.g., a group of arms, is
chosen, and the decision maker receives a direct reward
resulting from her strategy and also receives a \emph{side bonus} (either observation or reward)
resulting from that strategy for each arm's neighbors.

In this study, we take as input a relation graph $G$ that represents the correlation among $K$ arms.
In the standard setting, pulling an arm $i$ gets reward and observation $X_{i,t}$, while in the networked combinatorial bandit problem with side bonus, one also gets side observation or even reward due to the similarity or potential influence among neighboring arms. We consider two types of side bonus in this work:
(1) Side-observation: by pulling arm $i$  at time $t$ one gains the direct reward associated with $i$ and also observes the reward of her neighboring arms. Such
side-observation \cite{buccapatnam2013multi} is made possible in settings of on-line social networks where friends can provide their feedback on shared content, therefore if we promote a product to a user, we can also collect feedback from her friends on that product; (2) Side-reward:  in many practical applications such as recommendation in social networks, pulling an arm $i$ not only yields side observation on neighbors, but also receives extra rewards. That is by pulling arm $i$ one gains the reward associated with $i$ together with her neighboring arms directly. This setting is motivated by the observation that users are usually influenced by her friends when making purchasing decisions. \cite{myers2012information}.

Despite of many existing results on MAB problems against unknown stochastic environment \cite{anantharam1987asymptotically,kalathil2012decentralized,tekin2012online,audibert2009minimax,buccapatnam2013multi}, their adopted formulations  do not fit those applications that involve either side bonus or exponentially large number of candidate strategies.
There are several challenges facing our new study.
First of all, under combinatorial setting, the number of candidate strategies could be exponentially large,  if one simply treats each strategy as an arm, the resulting regret bound is exponential in the number of variables or arms.
Traditional MAB assumes that all the arms are independent, which is inappropriate in our setting. In the presence of side bonus, how to appropriately leverage additional information in order to gain higher rewards is another challenge.
To this end, we explore a more general formulation for \emph{networked combinatorial bandit problems} under four scenarios, namely, single/combinatorial play with side observation, single/combinatorial play with side reward.
The objective is to minimize the upper bound of regret (or maximize the total reward) over time.


The contributions of this paper are listed as follows:
\begin{itemize}
\item
For Single-play with Side Observation case,  we present the first distribution-free learning (DFL) policy, whose  time and space complexity
are bounded by $O(K)$. Our policy achieves zero regret that does not depend on  $\Delta_{\min}$, the minimum distance between the best static strategy and any other strategy.
\item For Combinatorial-play with Side Observation case, we present a learning policy with zero regret. Compared with traditional MAB problem without side bonus, we reduce the regret bound significantly.
\item For Single-play with Side Rewards case, we develop a distribution-free zero regret learning policy. We theoretically show that this scheme converges faster than any existing method.
\item For Combinatorial-play with Side Rewards case, by assuming that the combinatorial problem at each decision point can be solved optimally, we present the first distribution-free zero  regret policy.

\end{itemize}
We evaluate our proposed learning policy through extensive simulations and simulation results validate the effectiveness of our schemes.

The remainder of this paper is organized as follows.
We first give a
formal description of networked combinatorial multi-armed bandits
problem in Section~\ref{sec-model}. We study Single-play with Side Observation case in Section~\ref{sec-single-wo-reward}. In Section~\ref{sec-com-wo-reward}, we study Combinatorial-play with Side Observation case. Single-play with Side Rewards case has been discussed in Section~\ref{sec-single-wt-reward}. In Section \ref{sec-com-wt-reward}, we study Combinatorial-play with Side Rewards case. We evaluate our policies via extensive simulations in Section~\ref{sec-simu}. We review related works in Section~\ref{sec-review}. We conclude this paper, and discuss limitations as well as future works in Section~\ref{sec-con}.  Most notations used in this paper are summarized in Table \ref{notations}.

\section{Models and Problem Formulation}
\label{sec-model}

In the standard MAB problem, a $K$-armed bandit problem is defined by $K$ distributions $\mathcal{P}_1,\dots,\mathcal{P}_K$, each arm with respective means $\mu_1,\dots,\mu_K$.
When the decision maker pulls arm $i$ at time $t$, she receives a reward $X_{i,t}$.
We assume all rewards
    $\{
       X_{i,t}, i\in [1,K], t \geq 1
    \}$
are independent, and all $\{ \mathcal{P}_i\}$ have support in $[0,1]$.
Let $i=1$ denote the optimal arm, and $\Delta_i=\mu_1 - \mu_i$ be the difference between the best arm and arm $i$.

The relation graph $G=(V,E)$ over the $K$ arms describes the correlations among them, where an undirected link $e(i,j) \in E$  indicates the correlation between two neighboring arms $i$ and $j$.  In the standard setting, pulling an arm $i$ gets reward and observation $X_{i,t}$, while in the networked combinatorial bandit problem with side bonus, one also gets side observation or even reward from neighboring arms due to the similarity or potential influence among them. Let $N(i)$ denote the set of neighboring arms of arm $i$ and $N_i = \{i\}\cup N(i)$.
In this work, we consider two types of side bonus:
\begin{itemize}
\item \emph{Side observation}: by pulling arm $i$  at time $t$ one gains the reward $X_{i,t}$ associated with $i$ and also observes the reward $X_{j,t}$ of $i$'s neighboring arm $j\in N_i$. This is motivated by many real applications, for example, in today's online social network, friends can provide their feedback on shared content, therefore if we promote a product to one user, we can also collect feedback from her friends on that product;
\item \emph{Side reward}:  by  pulling an arm $i$ not only yields side observation on neighbors, but also receives  rewards from them, i.e., the total rewards would be $\sum_{ j \in N_i}  X_{j,t}$. This setting is motivated by the observation that in many practical applications such as recommendation in social networks, users are usually influenced by her friends when making purchasing decisions.
\end{itemize}

Upon the number of arms pulled at each time slot, we will study single-play case and combinatorial-play case.
\begin{itemize}
\item In the \emph{single-play case}, the decision maker selects one arm at each time slot, \eg, traditional MAB problem belongs to this category;
\item In the \emph{combinatorial-play case}, the decision maker requires to select a combination of $M (M \leq K)$  arms that satisfies given constraints.  One such example is online advertising, assume an advertiser can only place  up to $m$ advertisements on his website, he repeatedly selects a set of $m$ advertisements, observes the click-through-rate, with the goal of maximizing the average click-through-rate. This problem can be formulated as a combinatorial MAB problem where each arm represents one advertisement, subject to the constraint that one can play at most $m$ arms at each time slot.
In the combinatorial case,
at each time slot $t$, an $M$-dimensional \emph{strategy} vector
$\mathbf{s}_x$ is selected under some
\emph{policy} from the \emph{feasible strategy set} $F$. By feasible
we mean that each strategy satisfies the underlying constraints
imposed to $F$. 
We use $x=1,\dots, |F|$ to index strategies of feasible set $F$ in the
decreasing order of average reward  $\lambda_x$,
\eg, $\mathbf{s}_1$ has the largest average reward.
Note that a strategy may consist of less than $M$ random
variables, as long as it satisfies the given constraints.
We then set $i=0$ for any empty entry $i$.
\end{itemize}

In either case, the objective is to minimize long-term \emph{regret } after $n$ time slots, defined by cumulative difference between the received reward and the optimal reward.
\begin{table}
\begin{center}
\caption{ Summary of notations}
\label{notations}
\begin{tabular}{c| c}
\hline
Variable & Meaning \\
  \hline
    $K$     & number of arms        \\
    $M$     & number of selected arms          \\
    $G $    & relation graph over the arms\\
    $X_{i,t}$ & observation/direct reward on arm $i$ at time $t$\\
    $\mu_i$   & mean of $X_{i,t}$ \\
    $N_i$     & set of neighboring arms of arm $i$\\
    $\Delta_{i}$ & the distance between the best strategy and strategy $i$ \\
    $B_{i,t}$ & side reward received by arm $i$ from $N_i$ \\
    $O_{i,t}$ & number of observation times on arm $i$ by time $t$         \\
    $O_{i,t}^b$ & number of update times on side rewards of arm $i$ by time $t$         \\
    $\overline{X}_{i,t}$ & time averaged value of observation on arm $i$ by time $t$     \\
    $H$ & vertex-induced subgraph of $G$ composed by arms with $\Delta_i \geq \delta_0$   \\
    $\mathcal{C}$ & clique cover  of $H$  \\
    $F$     &  feasible strategy (arm or com-arm) set         \\
    $R_{x,t}$  &  direct reward on com-arm $x$ at time t\\
    $\sigma_x$ & mean of $R_{x,t}$   \\
    $Y_x$ & set of neighboring arms of component arms in com-arm $x$ \\
    $N$     & maximum of $|Y_x|$ among all com-arms  \\
    $CB_{x,t} $  & combinatorial side reward received by com-arm $x$ from $Y_x$  \\
    $\Delta_{x}$ & the distance between the best strategy and strategy $x$ \\
    $\Delta_{\min}$ & minimum of $\Delta_{x}$ among all strategies \\
  \hline
  \end{tabular}
\end{center}
\end{table}

 Consequently, this leaves us four scenarios to investigate: Single-play with Side Observation, Combinatorial-play with Side Observation, Single-play with Side Reward, and Combinatorial-play with Side Reward.
We then describe the problem formulation for each case. We use $I_t$ to denote index of selected arm (resp. strategy) by the decision maker at time slot $t$, and subscript $1$ to denote the optimal arm (resp. strategy) in the four cases. We evaluate policies using regret, $\mathfrak{R}_n$, which is defined as the difference in the total expected reward (over $n$ rounds) between always
playing the optimal strategy and playing arms according to the policy. We say a policy achieves \emph{zero regret} if  the expected average regret over time approaches zero as time goes to infinity, \ie, $\mathfrak{R}_n/n \rightarrow 0$ as $n\rightarrow \infty$. 
\begin{enumerate}
  \item \emph{Single-play with Side Observation (SSO)}. In this case, the decision maker pulls an arm $i$, observes all $X_{j,t}$, $j\in N_i$, and gets a reward $X_{i,t}$.
      The regret by time slot $n$ is written as,
      \begin{equation}
         \mathfrak{R}_n = \sum_{t=1}^{n} \mu_1 - \sum_{t=1}^{n} X_{I_t,t}.
      \end{equation}
      Here $I_t$ denotes the index of arm played at $t$.

  \item \emph{Combinatorial-play with Side Observation (CSO)}. Rather than pulling a single arm, the decision maker pulls a set of arms, $\mathbf{s}_{I_t}$, receives a reward \[R_{I_t,t} = \sum_{i \in \mathbf{s}_{I_t}} X_{i,t}\] and also observes reward $X_{j,t}$ for each neighboring arm  $j\in Y_{I_t}$, where  $Y_{I_t} = \cup_{i\in \mathbf{s}_{I_t}} N_i$ is the set of neighboring arms for selected strategy $I_t$. Therefore, let $\lambda_1$ denote the expected reward from the optimal strategy, the regret is defined as
       \begin{equation}
        \mathfrak{R}_n = \sum_{t=1}^{n} \lambda_1 - \sum_{t=1}^{n} R_{I_t,t}.
      \end{equation}
  \item \emph{Single-play with Side Rewards (SSR)}. When pulling an arm $i$, it yields a total reward
      \[B_{i,t} =  \sum_{ j \in N_i} X_{j,t}\]
      Therefore, the best arm shall be the one with the maximum expected total reward. Let $u_i = \sum_{j\in N_i} \mu_j$ denote the mean of reward for arm $i$, and $u_1$ the maximum reward. The regret is
      \begin{equation}
         \mathfrak{R}_n = \sum_{t=1}^{n} u_1 - \sum_{t=1}^{n} B_{I_t,t}.
      \end{equation}
      Note here, the optimal arm may differ from the optimal arm under single-play with side observation.
  \item \emph{Combinatorial-play Side Rewards (CSR)}. Different from combinatorial-play with side observation, the decision maker directly obtains the rewards from all neighboring arms.
      That is,  the totally received reward includes direct reward by strategy $x$ and side reward by its neighbors.
      Let $Y_x = \cup_{i\in \mathbf{s}_x} N_{i}$ be the set of neighboring arms for strategy $x$,
      and $\sigma_{x} = \sum_{i \in Y_{x}} \mu_i $ be the expected reward of $\mathbf{s}_x$.
       The combinatorial reward at time slot $t$ is written as $CB_{I_t,t} = \sum_{i \in Y_{I_t}} X_{i,t} $. We define the regret as
      \begin{equation}
        \mathfrak{R}_n = \sum_{t=1}^{n} \sigma_1 - \sum_{t=1}^{n} CB_{I_t,t}.
      \end{equation}
\end{enumerate}

\section{Single-play  with side observation}
\label{sec-single-wo-reward}

We start with the case of
Single-play with Side Observation.
In this case, the decision maker
learns to select an arm (resp. strategy) with
maximum reward, meanwhile
observes side information of
its neighbors defined in relation graph.
Our proposed policy,
which is the first distribution free learning policy
for SSO reffered to as DFL-SSO,
is shown  in Algorithm~\ref{alg-swo}.
As shown in Line 2-5,
the decision maker updates
all neighbors' side information,
i.e., number of observation up
to current time, and time-averaged
reward.
The key idea behind the algorithm
is that side-observation
potentially reduces the regret
as the decision maker can explore
more without pain, thus gain more
history information to exploit.

To theoretically analyze the benefit of side observation,
we novelly leverage the technique of graph partition
and clique cover. 
The basic idea in standard analysis of regret bound with
side observation in distribution-dependent case is
to use clique cover of relation graph,
and use the arm with maximum $\Delta_i$ inside each cilque
to represent the clique for analysis.
While  standard proof of distribution-free
regret bound is to divide the arms into two sets via
a threshold $\Delta_{c_0}$ on $\Delta_i$, and then respectively
analyze the bounds of the two sets of arms.
Therefore, to obtain a distribution-free result,
we cannot directly use the arm with maximum $\Delta_i$
inside a clique for representation to prove
distribution-free regret bound,
as the arms with $\Delta_i$ smaller than $\Delta_{c_0}$
are distributed inside cliques.
To address this issue, we first partition the relation
graph $G$ using the predefined threshold,
and then mainly analyze the benefit of side observation
in one vertex-induced subgraph $H$
for arms having $\Delta_i$ above $\Delta_{c_0}$.
In the subgraph $H$, it is then possible
to analyze the distribution-free 
regret bound using the technique of clique cover.

\begin{figure}[t]
    \centering
    \includegraphics[width=3in]{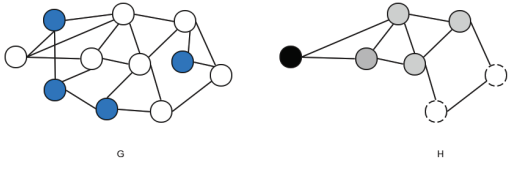}
    \caption{Graph partition: $G$ is relation graph, and $H$ is vertex-induced graph that is covered by $3$ cliques }
                  \label{fig-pt}
\end{figure}

  Theorem~\ref{theo-swo}
  quantifies the benefit
  brought about by it,
  where it shows that the more
  side observation (e.g., smaller
  clique number) is,
  the smaller the upper bound
  of regret is.

\begin{algorithm}[ht]
\caption{Distribution-Free Learning policy for single-play with side observation (DFL-SSO)}
  \begin{algorithmic}[1]
  \label{alg-swo}
   \STATE For each time slot $t=0,1,\dots,n$ \\*
    Select an arm $i$ by maximizing
        \begin{equation}
        \label{eqn-swo}
        \overline{X}_{i,t} + \sqrt{ \frac{ \log{( t/(KO_{i,t}))}} {O_{i,t}}}
        \end{equation}
    to pull
    \STATE for $k \in N_i$ do
        \STATE $O_{k,t+1}  \leftarrow  O_{k,t} +1 $
        \STATE $\overline{X}_{k,t+1} \leftarrow X_{k,t}/O_{k,t}
                + (1- 1/O_{k,t}) \overline{X}_{k,t}$
    \STATE end for
    \STATE end for
  \end{algorithmic}
\end{algorithm}

\begin{theorem}
\label{theo-swo}
    The expected regret of Algorithm~\ref{alg-swo} after $n$ time slots is bounded by
\begin{eqnarray}
    \mathcal{R}_n &\leq& 15.94 \sqrt{nK} +
    0.74 \mathcal{C} \sqrt{n/K},
\end{eqnarray}
    where $\mathcal{C}$ is clique cover of  vertex-induced subgraph $H$ with arms of $\Delta_i$ above threshold $\delta_0$ in relation graph $G$.
\end{theorem}

\begin{IEEEproof}
	The proof is based on our novel combination of graph partition and clique cover.
	We first  partition relation graph to rewrite regret in terms of cliques, and then 
	mainly tighten the upper bound by analyzing regret of cliques.

    \textbf{1. Partition relation graph  and rewrite regret of subgraph $H$ in terms of cliques.}
    
    We order the arms in an increasing order of $\Delta_i$.
    We use $\Delta_{c_0} \leq \delta_0 = \alpha \sqrt{K/n} \leq \Delta_{c_0+1}$  to split the $K$ arms into two disjoint sets, one set $K_1$ with $\Delta_x \leq \Delta_{c_0}$ and the other set $K_2$ with $\Delta_x > \Delta_{c_0}$ (We will set the value of $\alpha$ in later analysis).
    Let $c_0$ be the smallest index of arm satisfying
    $\Delta_k \leq \Delta_{c_0}$.
    We remove all arms in $K_1$ from the relation graph $G$, as well as adjacent edges to nodes in $K_1$.
    In this way, we get a subgraph $H$ of $G$, over arms in $K_2$. The regret satisfies,
    \begin{equation}
        \mathfrak{R}(n) \leq  n \Delta_{c_0} + \mathfrak{R}_H(n),
    \end{equation}
    where $\mathfrak{R}_H(n)$ is regret generated by selecting
    suboptimal arms in $K_2$.

    Consider a clique covering $\mathcal{C} $ of $H$, i.e., a set  of cliques such that
        each $c \in \mathcal{C} $
    is a clique and $V = \cup_{c \in \mathcal{C} } c$.
    We define the clique regret $\mathfrak{R}_c(n)$ for any
    $c \in \mathcal{C}$ by
    \begin{equation}
        \mathfrak{R}_c(n) = \sum_{t<n}\sum_{i\in c} \Delta_i \mathbf{1}\{I_t = i\}.
    \end{equation}

    Since the set of cliques covers the whole graph $H$,
    we have
    \begin{equation}
        \mathfrak{R}_{H}(n) \leq \sum_{c \in \mathcal{C}}  R_c (n).
    \end{equation}
    
    We give an illustration of the partition process in Fig.~\ref{fig-pt}, where the relation graph $G$
contains one small set of blue nodes representing $K_1$ with $\Delta_i$ below $\Delta_{c_0}$,
and the other large set of white nodes denoting $K_2$ with
$\Delta_i$ above $\Delta_{c_0}$.
The vertex-induced subgraph $H$ of $K_2$ is covered
by a minimum of $3$ cliques, respectively marked by black, gray and dash lines.

\textbf{2. Regret analysis for regret of subgraph $H$ }

In the rest part, we focus on proving upper bound of regret $\mathfrak{R}_{H}(n)$.
    Let $\Delta_c = \max_{i \in c} \Delta_i$,
    and $T_c(t) = \sum_{i \in c} T_i(t)$ denote the number of times (any arm in) clique $c$ has been played up to time $t$, where $T_i(t)$ is the number of times arm $i$
       has been selected up to time $t$.
    Similarly, we suppose that cliques are ordered in the increasing order of $\Delta_c$.
Let $v_j = \mu_1 - \frac{\Delta_j}{2}$ for cliques in $K_2$,
$c_0 \leq j \leq K $,
and $v_{c_0} = \mu_1 - \frac{\Delta_{c_0}}{2}$.
Let $z_{c_0} = + \infty$ and $\Delta_{K+1}= + \infty$.
For better description, we use $c_0$ to denote the case of $c=0$.

As every arms in a clique $c$ must be observed for the same number of times,
then for each clique and $l_0\geq 0$, we have
\begin{eqnarray}
   \mathfrak{R}_c =\sum_{i\in c} \Delta_i T_i(n) \leq l_0 \max_{i\in c} \Delta_i + \sum_{i\in \mathcal{C}} \sum_{l = l_0}^{\infty}\mathbf{1}\{I_t = i, t \geq l_0\}
\end{eqnarray}
Meanwhile,
\begin{eqnarray}
   \mathfrak{R}_{H}(n) =  \sum_{c\in K}\mathfrak{R}_c
   = \sum_{c \in \mathcal{C}} l_0 \Delta_c + \sum_{i=1}^{K} \Delta_i T_i'(n),
\end{eqnarray}
Where $T_i'(n)$ denotes the number of arm $i$ played after $t=l_0$, and we refer to the second term as $\mathfrak{R}_H'$

Define
\begin{equation}
    W = \min_{1\leq t \leq n } W_{1,t},
\end{equation}
and
\begin{equation}
    U_{j,i}  = \mathbf{1}_{W\in[v_{j+1},v_j)} \Delta_i T_i'(n).
\end{equation}

We have the following for $\mathfrak{R}_H'(n)$,
\begin{eqnarray}
    \mathfrak{R}_H'(n)  &=& \sum_{i=c_0}^{K}\Delta_i T_i'(n)
    \\
    &=& \sum_{j=c_0}^{K}\sum_{i=1}^{j}  U_{j,i}  + \sum_{j=c_0}^{K} \sum_{i=j+1}^{\mathcal{C}} U_{j,i}.
    \label{eqn-rh}
\end{eqnarray}


For the first term of Equation~(\ref{eqn-rh}),
    we have:
\begin{eqnarray}
    \sum_{j=c_0}^{K}\sum_{i=1}^{j}  U_{j,i}
  &\leq&
    \sum_{j=c_0}^{K} \mathbf{1}_{W \in [v_{j+1},v_j)} n \Delta_j  \\
  &=&
    n \Delta_{c_0} + n \sum_{c=1}^{\mathcal{C}} \mathbf{1}_{W \leq v_{c}} (\Delta_c - \Delta_{c-1}).
\end{eqnarray}
We have the first equation as $\Delta_j \geq \Delta_i$ and $T_i \leq n$.

To bound the second term of Equation~(\ref{eqn-rh}),
we record
\begin{equation}
    \tau_i = \{\min{t: W_{i,t} < v_i}\}
 \end{equation}after $l_0$.
To pull a suboptimal arm $i$ at $t$,
one must have $W_{i,t} > W_{1,t} \geq  W$.
By Algorithm~\ref{alg-swo}, we have
$\{W\geq v_i\} \subset\{T_i'(n)\leq \tau_i\}$,
since once we have pulled $\tau_i$ times arm $i$ its index will always be lower than the index of arm 1.

Therefore, we have
\begin{eqnarray}
     \mathfrak{R}(n)&\leq&  2 n \Delta_{c_0} + \sum_{c \in \mathcal{C}} l_0 \Delta_c
     +
      \sum_{i=1}^{K}\Delta_i \mathbf{E}(\tau_i|t>l_0)
     \nonumber \\
   &&
     + n \sum_{c=1}^{\mathcal{C}} \mathbf{1}_{W <v_c}(\Delta_c - \Delta_{c-1}).
\end{eqnarray}

For any $l_0 >0$,
\begin{eqnarray}
  && {\Delta_i}\mathbf{E}(\tau_i | \tau_i >l_0) \\
    &\leq&
        \sum_{l=l_0}^{+\infty}\mathbf{P}(\tau_i \geq l)      \label{eqn-tau}  \nonumber    \\
    &=&   \sum_{l=l_0}^{+\infty}\mathbf{P}(\forall t\leq l,W_{i,t} > v_i )
   \nonumber \\
    &\leq&  \sum_{l=l_0}^{+\infty}\mathbf{P}
                    \biggl(
                            \overline{X}_{i,l}-\mu_i \geq
                            \frac{\Delta_i}{2}-\sqrt{\frac{\log_+(n/Kl)}{l}}
                    \biggl) \nonumber
                    \\
\end{eqnarray}

Let $l_0 = 8 \log{(\frac{n}{K}\Delta_i^2})/\Delta_i^2$. For $l\geq l_0$,  we have
\begin{eqnarray}
       \log_+(t/(Kl))
   \leq \log_+ (n/(Kl_0))
   \leq (\frac{n}{K} \times \frac{\Delta_i^2}{8}) \\
   \leq \frac{l_0\Delta_i^2}{8} \leq \frac{l \Delta_i^2}{8}.
\end{eqnarray}
Therefor, we have
\begin{equation}
 \frac{\Delta_i}{2}-
  \sqrt{\frac{\log_+(n/Kl)}{l}}
  \geq \frac{\Delta_i}{2}- \frac{\Delta_i}{\sqrt{8}} = a\Delta_i
\end{equation}
with $a=\frac{1}{2}-\frac{1}{\sqrt{8}}$,

\begin{eqnarray}
    \Delta_c l_0
        &\leq&
            8 \log{(\frac{n}{K}\Delta_i^2})/\Delta_i
         \leq
         \frac{2}{e}\sqrt{n/K}
\end{eqnarray}

To
bound (\ref{eqn-tau}) using Hoeffding Bound, i.e.,
\begin{eqnarray}
  \mathbf{E}\{\tau_i| t>l_0 \}
  &\leq&
          \sum_{l=l_0}^{+\infty}
   \mathbf{P}(\overline{X}_{i,l}-\mu_i \geq a\Delta_i ) \\
  &\leq&  \sum_{l=l_0}^{+\infty} \exp{(-2l(a\Delta_i)^2)}  \\
  &=&  \sum_{l=l_0}^{+\infty} \frac{1-2 l_0(a\Delta_i)^2}{1-\exp(-2(a\Delta_i)^2)} \\
  &\leq& \frac{1}{1-\exp(-2(a\Delta_i)^2)}           \\
  &\leq& \frac{1}{(2a\Delta_i)^2 -(-2(a\Delta_i)^2)} \\
  &=& \frac{1}{2a\Delta_i^2(1-a^2)}.
\end{eqnarray}

Then we have
\begin{eqnarray}
    \Delta_i\mathbf{E}\{\tau_i | t>l_0\}
        &\leq&
            8 \log{(\frac{n}{K}\Delta_i^2})/\Delta_i
          + \frac{1}{2a\Delta_i(1-a^2)} \nonumber \\
        &\leq&
         \frac{2}{e}\sqrt{n/K} + \frac{\alpha^{-1}}{2a(1-a^2)}\sqrt{n/K}.
\end{eqnarray}


Now we prove to bound
$n\sum_{c=0}^{\mathcal{C}}
 \mathbf{P} (W \leq v_{c}) (\Delta_c - \Delta_{c-1}) $.
Recall that
  $\Delta_{c_0} \leq \delta_0 \leq \Delta_{c_0+1} $,
  and let $\delta_{c_0}$ be $\Delta_{c=0}$.
Taking $\mathbf{P}(W \leq \mu_1 - \frac{\Delta_c}{2})$ as an nonincreasing function of $\Delta_c$, we have
\begin{eqnarray}
    \sum_{c=1}^{\mathcal{C}} \mathbf{P}(W \leq v_c) (\Delta_c- \Delta_{c-1})  \nonumber \\
    \leq \delta_0 - \Delta_{c_0} + \int_{\delta_0}{1} \mathbf{P}(W \leq \mu_1 - \frac{u}{2})du.
\end{eqnarray}
For a fixed $u\in[\delta_0,1]$ and $f(u) = 8\log(\sqrt{n/K}u)/u^2$, we have
\begin{eqnarray}
    && \mathbf{P}(W \leq \mu_1 - \frac{u}{2}) \nonumber \\
    &=& \mathbf{P}
        \biggl(
                \exists 1\leq l \leq n: \overline{X}_{1,l} + \sqrt{\frac{\log{(n/(Kl)})}{l}}
                < \mu_1 - \frac{u}{2}
         \biggl)                                \nonumber \\
    &\leq&
        \mathbf{P}
        \biggl(
                \exists 1\leq l \leq f(u):
                 \mu_1 - \overline{X}_{1,l} > \sqrt{\frac{\log{(n/(Kl)})}{l}}
         \biggl)  \nonumber \\
    &~&   + \mathbf{P}
        \biggl(
                \exists 1\leq l \leq f(u):
                 \mu_1 - \overline{X}_{1,l} >
                 \frac{u}{2}
         \biggl)
         \label{eqn-peeling}
\end{eqnarray}
Let $P_1$ denote the first term of (\ref{eqn-peeling}), using
the form of
$\frac{1}{2^{m+1}}f(u) \leq l \leq \frac{1}{2^{m}}f(u)$,
we have
\begin{eqnarray}
    P_1 &\leq& \sum_{m=1}^{\infty} \mathbf{P} \biggl(
            \exists \frac{1}{2^{m+1}}f(u) \leq l \leq \frac{1}{2^{m}}f(u): \nonumber\\
         &&  l( \mu_1 - \overline{X}_{m,l} ) >
            \sqrt{
                \frac{f(u)}{2^{m+1}} \log(\frac{n2^m}{Kf(u)})
            }
        \biggl)   \nonumber \\
        &\leq& \sum_{m=1}^{\infty} \exp \biggl(
            -2\frac{f(u)2^{-(m+1)}\log(\frac{n2^{m}}{\mathcal{C}f(u)})}{f(u)2^{-m}}
        \biggl) \nonumber \\
       & =& 2\frac{Kf(u)}{n}
\end{eqnarray}
Let $P_2$ denote the first term of (\ref{eqn-peeling}), using
the form of
$2^{m}f(u) \leq l \leq 2^{m+1}f(u)$,
we have similarly,
\begin{eqnarray}
    P_2 &\leq&
    \sum_{m=1}^{\infty} \mathbf{P} \biggl(
            \exists 2^{m}f(u) \leq l \leq 2^{m+1}f(u): \nonumber \\
         &&  l( \mu_1 - \overline{X}_{m,l} ) >
            \frac{lu}{2}
        \biggl) \nonumber \\
        &\leq&\sum_{m=0}^{\infty}
        \exp\biggl(
            -2\frac{(2^{m-1}f(u)u)^2}{f(u)2^{m+1}}
        \biggl)   \nonumber \\
        &\leq& \frac{1}{\exp(f(u)u^2/4)-1}  \nonumber \\
       &\leq& \frac{1}{nu^2/K-1}
\end{eqnarray}
The last inequality comes from $f(u)$ is upper bounded by $4n/(eK)$.

By taking integrity on $P_1$ and $P_2$, we respectively have
\begin{eqnarray}
n \int_{\delta_0}^{1} P_1 du &\leq& n \frac{2K}{n} \int_{\delta_0}^1 f(u)du \\
    &=&   n \frac{2K}{n}
            \biggl[
                    \frac{8\log(e\sqrt{n/K}u)}{u}
            \biggl]_{1}^{\delta_0} \nonumber\\
    &\leq& \frac{8\log(e\alpha)}{\alpha} \sqrt{nK},
\end{eqnarray}

and
\begin{eqnarray}
   n \int_{\delta_0}^{1}P_2 du \leq
   \frac{1}{2} \log \biggl(
    \frac{\alpha+1}{\alpha-1}
    \biggl)\sqrt{nK}.
\end{eqnarray}
Instantly we have
\begin{eqnarray}
   && n\sum_{c=0}^{\mathcal{C}}
 \mathbf{P} (W \leq v_{c}) (\Delta_c - \Delta_{c-1}) \nonumber \\
 &\leq&
 n(\delta_0 - \Delta_{c_0})
 +
 \biggl(
 \frac{8\log(e\alpha)}{\alpha}
 +
 \frac{1}{2} \log \biggl(
    \frac{\alpha+1}{\alpha-1}
    \biggl)
 \biggl) \sqrt{nK} \nonumber
\end{eqnarray}

Finally, we get the regret bounded by

\begin{eqnarray}
    \mathcal{R}_n &\leq&
    \sum_{c\in \mathcal{C}} \frac{2}{e}\sqrt{n/K}
    +
    \biggl(
    3\alpha+
    \frac{8\log(e\alpha)}{\alpha}
    +
    \frac{1}{2} \log \biggl(\frac{\alpha+1}{\alpha-1}\biggl)
    +   \nonumber \\
   &&
      \frac{\alpha^{-1}}{2a(1-a^2)}
          \biggl)
     \sqrt{nK}
\end{eqnarray}

Let $\alpha =e$, and we already have $a = \frac{1}{2}-\frac{1}{\sqrt{8}}$,
then
\begin{eqnarray}
    \mathcal{R}_n &\leq& 15.94 \sqrt{nK} +
    0.74 \mathcal{C} \sqrt{n/K}.
\end{eqnarray}
\end{IEEEproof}

\section{Combinatorial-play with side observation}
\label{sec-com-wo-reward}
In this section,
we consider combinatorial-play with side observation.
In this case,
an intuitively  extension is to
take each strategy as an arm ( we
name it \emph{com-arm}),
and then apply the algorithm for SSO
to solve the problem.
However, the key question is how to utilize
the side-observation on arms defined
in relation graph to
gain more observation on com-arms,
that is, how to define neighboring
com-arms.
To this end, we introduce
the concept of strategy relation
graph to model
the correlation among
com-arms, by which we convert
the problem of CSO to SSO.

\begin{figure}[t]
    \centering
    \includegraphics[width=3in]{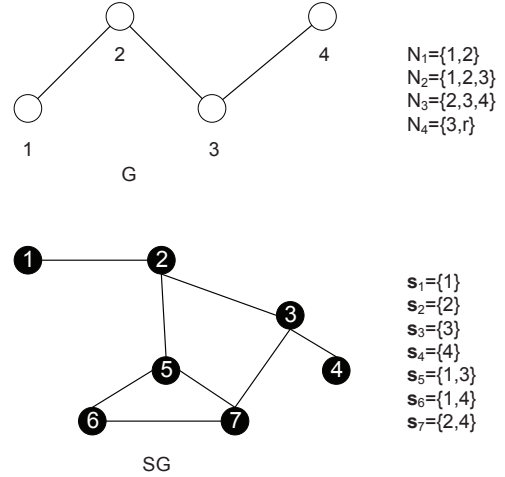}
    \caption{Convert combinatorial-play to single-play: constructing strategy relation graph $SG(F,L)$
    based on arm relation graph $G$ }
                  \label{fig-convert}
\end{figure}

The construction
process for strategy
relation graph is as follows.
We define strategy relation graph $SG
 (F, L)$ for strategies in $F$,
where $F$ is vertex set, and $L$ is edge set.
Each strategy $\mathbf{s}_x$ is denoted by a vertex,
and a link $\mathbf{l}=(\mathbf{s}_x,\mathbf{s}_y)$ in $L$ connects
two distinct vertexes $\mathbf{s}_x$ and $\mathbf{s}_y$
if $\mathbf{s}_y \in Y_x$ and vice versa.
The neighbor definition for strategies is natural
as once a strategy is played, the union of neighbors of arms in this strategy could be observed according to
neighbor definition for arms in $G$,
which surely reward of any strategy composed by these observed arms is
also observed.
We give an example in Fig.~\ref{fig-convert}.
There are $4$ arms in relation graph $G$,  indexed by
$i=1,2,3,4$.
The combinatorial MAB problem is to select a maximum weighted
independent set of arms where unknown bandit is weight.
As shown in Fig.~\ref{fig-convert}, the feasible strategy set
for this problem consists of $7$ feasible strategies, i.e.,
independent sets of arms in $G$:
\begin{eqnarray}
&&\mathbf{s}_1=\{1\},
    \cup_{i\in \mathbf{s}_1}N_i= \{1,2\}\nonumber \\
&&\mathbf{s}_2=\{2\},
    \cup_{i\in \mathbf{s}_2}N_i= \{1,2,3\}\nonumber \\
&&\mathbf{s}_3=\{3\},
    \cup_{i\in \mathbf{s}_3}N_i= \{2,3,4\}\nonumber \\
&&\mathbf{s}_4=\{4\},
    \cup_{i\in \mathbf{s}_4}N_i= \{3,4\} \nonumber \\
&&\mathbf{s}_5=\{1,3\},
    \cup_{i\in \mathbf{s}_5}N_i= \{1,2,3,4\}\nonumber \\
&&\mathbf{s}_6=\{1,4\},
    \cup_{i\in \mathbf{s}_6}N_i= \{1,2,3,4\}\nonumber \\
&&\mathbf{s}_7=\{2,4\},
    \cup_{i\in \mathbf{s}_7}N_i= \{1,2,3,4\} \nonumber
\end{eqnarray}
Taking $\mathbf{s}_2$ and $\mathbf{s}_5$ for illustration,
the component arms of $\mathbf{s}_2$, i.e., $\{2\}$, is a
subset of $\cup_{i\in \mathbf{s}_5}N_i= \{1,2,3,4\}$,
and the component arms of $\mathbf{s}_5$, i.e., $\{1,3\}$
is also a subset of $\cup_{i\in \mathbf{s}_2}N_i= \{1,2,3\}$.
Therefore, the two strategies are connected in the relation graph $SG$.

Consequently, we can convert the combinatorial-play MAB
with side observation to a single-MAB with side observation.
More specifically, taking each strategy as an arm, $SG(F,  L)$ is exactly a relation graph for com-arms in $F$.
The problem turns into a single-play MAB problem where at each time slot the decision maker  selects one com-arm from
$|F|$ ones to maximize her long-term reward.

\begin{algorithm}[ht]
\caption{Distribution-Free Learning policy for combinatorial-play with side observation (DFL-CSO)}
  \begin{algorithmic}[1]
  \label{alg-cwo}
   \STATE For each time slot $t=0,1,\dots,n$ \\*
    Select a com-arm $\mathbf{s}_{x}$  by maximizing
        \begin{equation}
        \label{eqn-cwo}
        \overline{R}_{x,t} + \sqrt{ \frac{ \log{( t/(KO_{x,t}))}} {O_{x,t}}}
        \end{equation}
    to pull
    \STATE UPDATE: for $y \in  N_x$ do
        \STATE $O_{y,t+1}  \leftarrow  O_{y,t} +1 $
        \STATE $\overline{R}_{y,t+1} \leftarrow R_{y,t}/O_{y,t}
                + (1- 1/O_{y,t}) \overline{R}_{y,t}$
    \STATE end for
    \STATE end for
  \end{algorithmic}
\end{algorithm}

%
%

The algorithm is shown in Algorithm~\ref{alg-cwo}, and we derive the regret bound below directly.
\begin{theorem}
    The expected regret of Algorithm~\ref{alg-cwo} after $n$ time slots is bounded by
\begin{eqnarray}
    \mathcal{R}_n &\leq& 15.94 \sqrt{n |F|} +
    0.74 \mathcal{C} \sqrt{n/|F|}.
\end{eqnarray}
\end{theorem}
In the traditional distribution-free MAB by taking each com-arm as an unknown variable\cite{audibert2009minimax},
the regret bound would be $49\sqrt{n|F|}$.
Our theoretical result significantly reduces the regret and tightens the bound.

\section{Single-play with side rewards}
\label{sec-single-wt-reward}
Though the single-play MAB with side reward
have the same observation as
the single-play MAB with side observation,
the distinction on reward function
makes the problem different.
In the case of SSR,
the reward function
is side reward of the selected arm $I_t$,
instead of its direct reward.
Here we  treat the side reward of each
arm as a new unknown random variable,
i.e., we require to learn $B_{i,t}$
that is a combination of all direct rewards in $N_i$.
As direct rewards of arms in $N_i$ are observed asynchronously, 
we cannot update the observation on $B_{i,t}$
as the way in SSO where 
observation is
 symmetric between two neighboring nodes. 
The trick is  
 updating the number of observation on $B_{i,t}$
 only when direct rewards of all arm in $N_i$
are renewed.
We use $O_{i,t}^b$ to denote this quantity to differ from
$O_{i,t}$ which denotes the number of direct reward is observed.
Therefore, whenever an arm is played or
its neighbor is played, the number of observation on side reward $O_{i,t}^b$ 
can be updated only when the
least frequently observed arm in $N_i$ is updated.
That is,
\begin{eqnarray}
\label{eqn-updt}
O_{i,t}^b =
    \begin{cases}
        O_{i,t-1}^b +1
                & \mbox{if $\min_{j\in N_i} O_{j,t}$ is updated}
        \\
        O_{i,t}^b &\mbox{Otherwise.}
    \end{cases}
\end{eqnarray}

The algorithm for
single-play MAB with side reward is
summarized in Algorithm~\ref{alg-swb}
where we directly use
side reward $B_{i,t}$ as observation,
and update   $O_{i,t}^b$
according to (\ref{eqn-updt}).
The regret bound of our proposed algorithm
is presented in Theorem~\ref{theorem-ssr}.

\begin{algorithm}[ht]
\caption{Distribution-Free Learning policy for single-play with side reward (DFL-SSR)}
\label{alg-swb}
  \begin{algorithmic}[1]
   \STATE For each time slot $t=0,1,\dots,n$ \\*
    Select an arm $i$ by maximizing
        \begin{equation}
        \label{eqn-swb}
      \overline{B}_{i,t} + \sqrt{ \frac{ \log{( t/(K O_{i,t}^{b}))}} {O_{i,t}^{b}}}
        \end{equation}
    to pull
    \STATE for $k \in N_i$ do
        \STATE $O_{k,t+1}  \leftarrow  O_{k,t} +1 $
        \STATE if  $\min_{j\in N_k} O_{j,t}$ is updated
            \STATE $O_{k,t+1}^b = O_{k,t}^b+1$
            \STATE $\overline{B}_{k,t+1} =
                \overline{B}_{k,t}/O_{k,t}^b+ (1-1/O_{k,t}^b)\overline{B}_{k,t}$
        \STATE end if
    \STATE end for
    \STATE end for
  \end{algorithmic}
\end{algorithm}

\begin{theorem}
	\label{theorem-ssr}
    The expected regret of Algorithm~\ref{alg-swb} after $n$ time slots is bounded by
\begin{eqnarray}
    \label{rgt-swb}
    \mathcal{R}_n &\leq& 49K \sqrt{nK}
\end{eqnarray}
\end{theorem}
\begin{IEEEproof}
In this case, $B_{i,t} \in [0,K]$,
which indicates that the range of received reward is scaled by $K$ at most. We normalize $B_{i,t} \in [0,1]$.
 Using the same techniques in proof of MOSS algorithm \cite{audibert2009minimax}, we get the normalized regret bound,
 and  then the regret bound in (\ref{rgt-swb})
 by scaling the normalized regret bound by $K$.
In  Algorithm~\ref{alg-swb}, the number of observation times on side reward should be no less than the scenario without side observation.
Therefore,  Algorithm~\ref{alg-swb} would convergence to the optimality faster than the MOSS algorithm without side observation.
\end{IEEEproof}

\section{Combinatorial-play with side rewards}
\label{sec-com-wt-reward}
\begin{figure*}[!pt]
    \centering
    \subfigure[Expected regret]
    {
            \includegraphics[ width=2.5in]{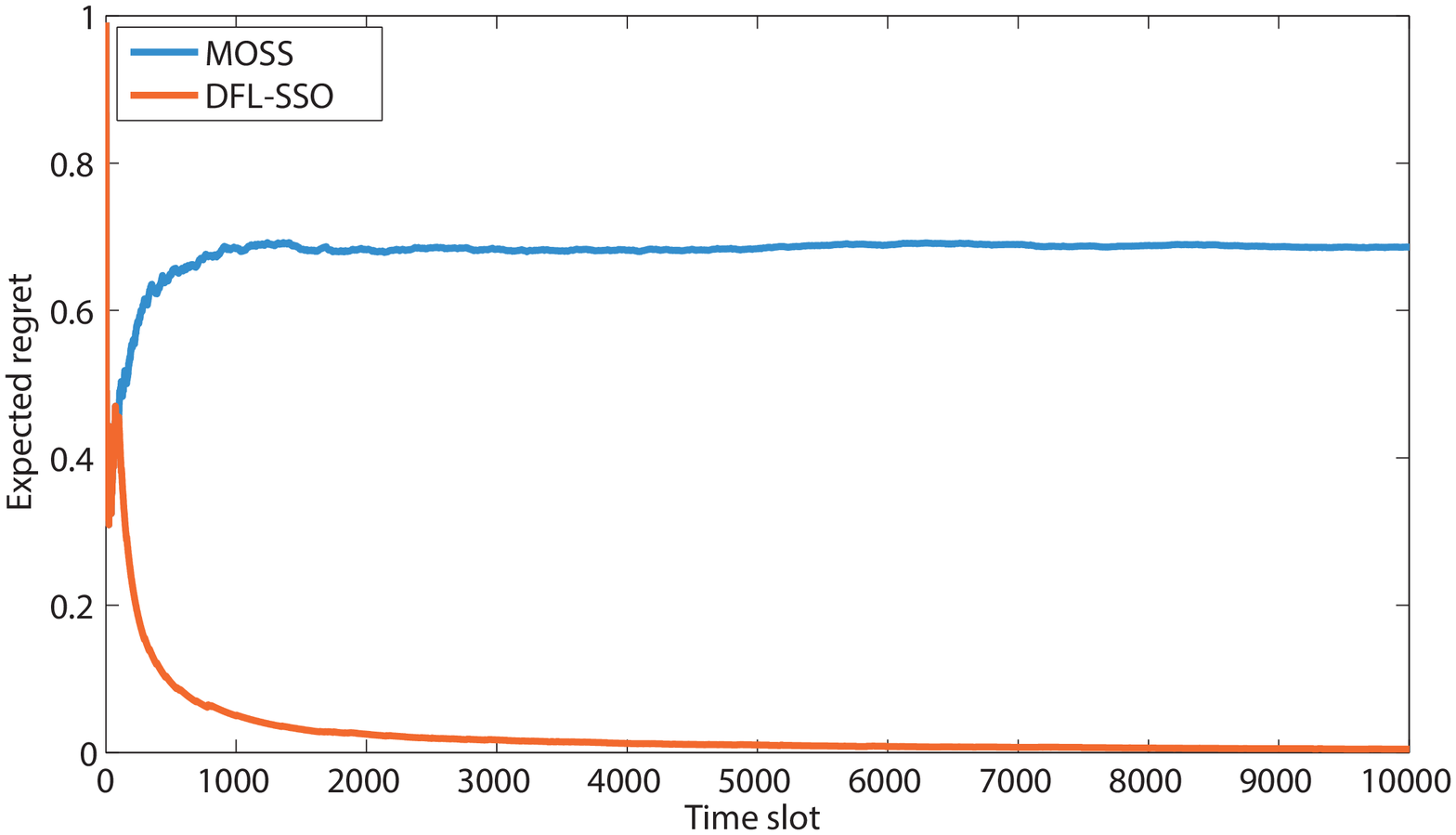}
                    \label{fig-swb-exp}
    }
        \subfigure[Accumulated regret]
    {
          \includegraphics[width=2.5in]{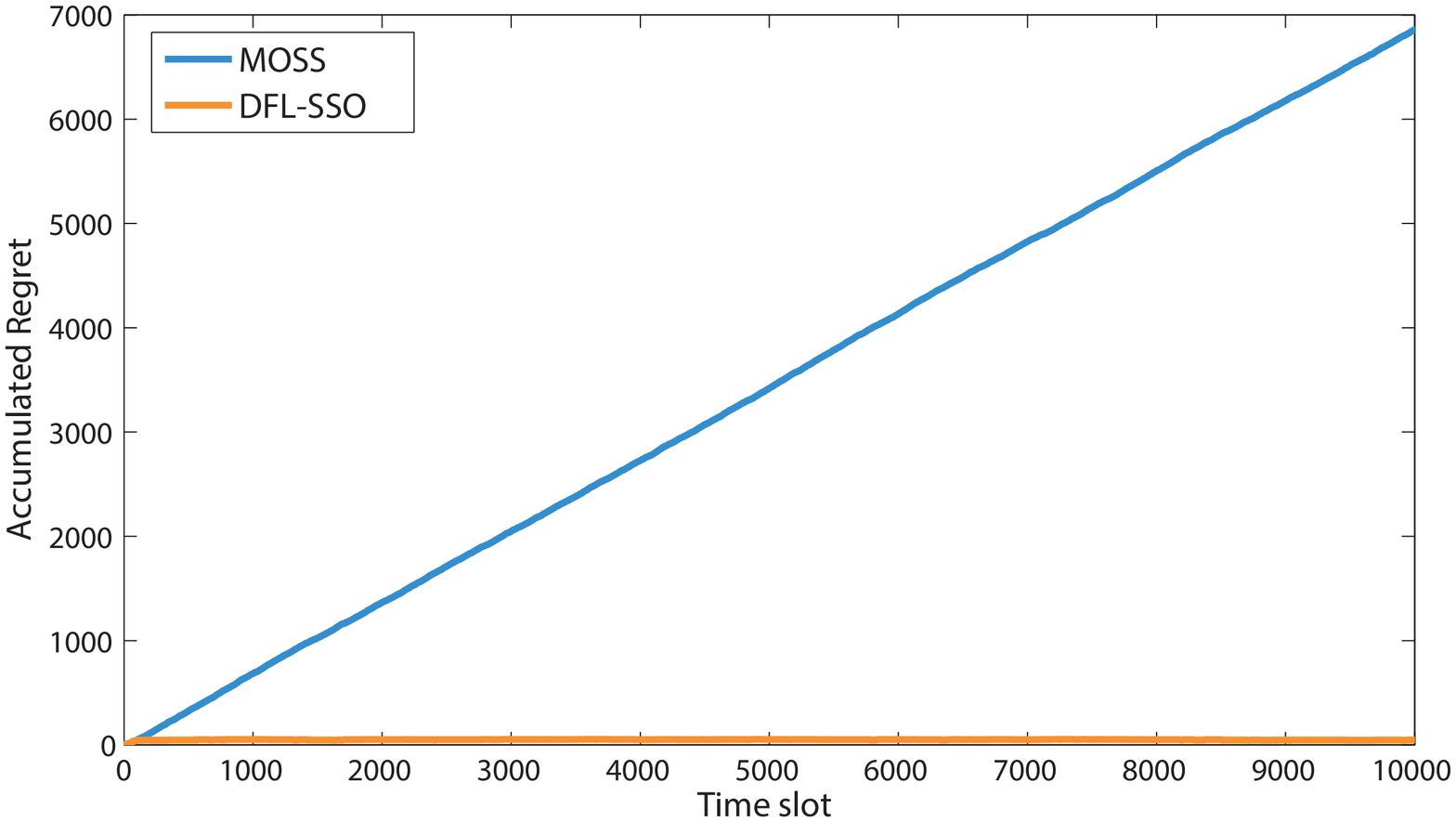}
                \label{fig-swb-acc}
                  \hfill
    }
  \caption{Comparison of regret: MOSS v.s. DFL-SSO}
  \label{fig-alg1}
\end{figure*}\begin{figure*}[ptb]
    \centering
    \subfigure[Sparse relation graph ]
    {
            \includegraphics[ width=2.5in]{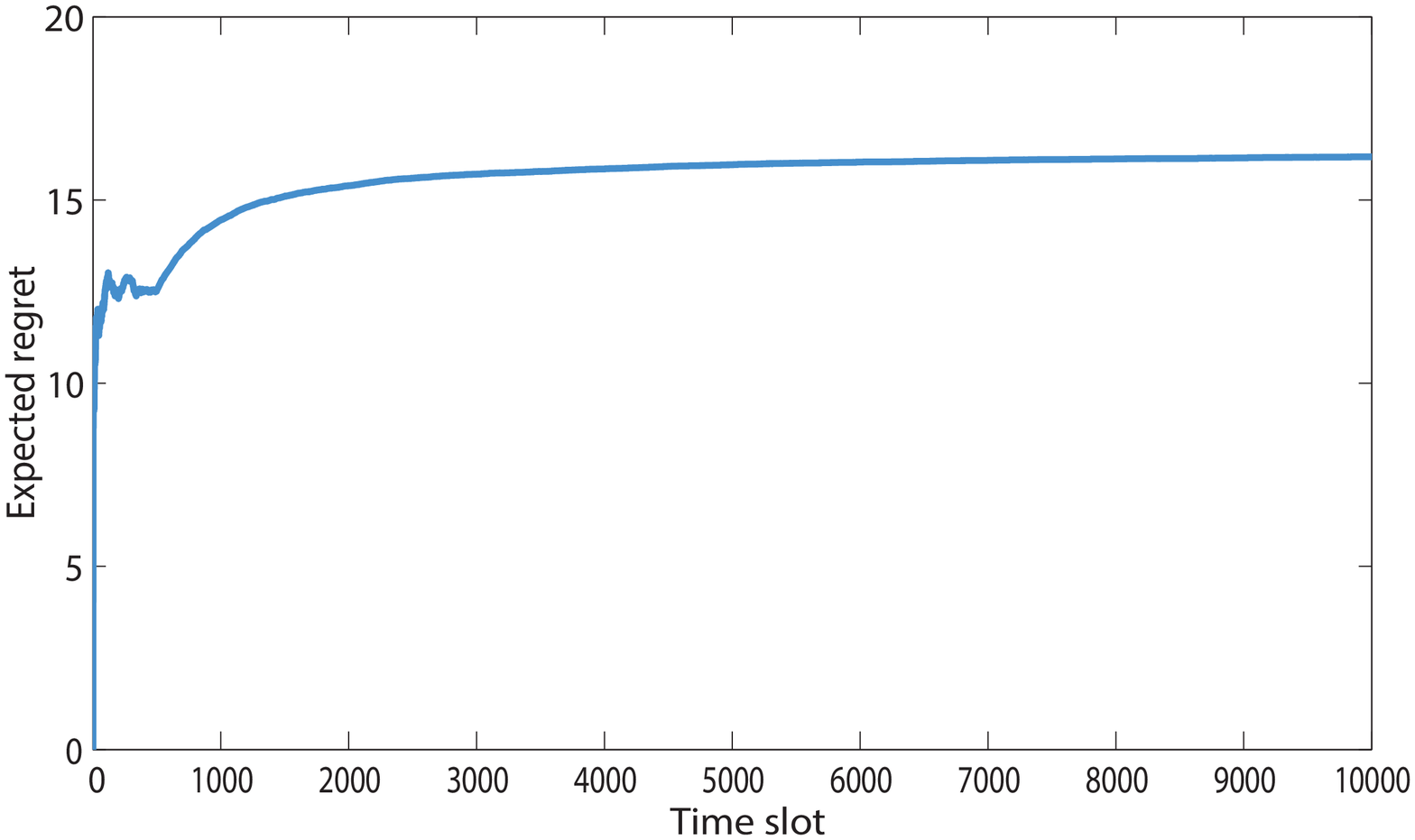}
                    \label{fig-alg2-03}
    }
        \subfigure[Dense relation graph ]
    {
          \includegraphics[width=2.5in]{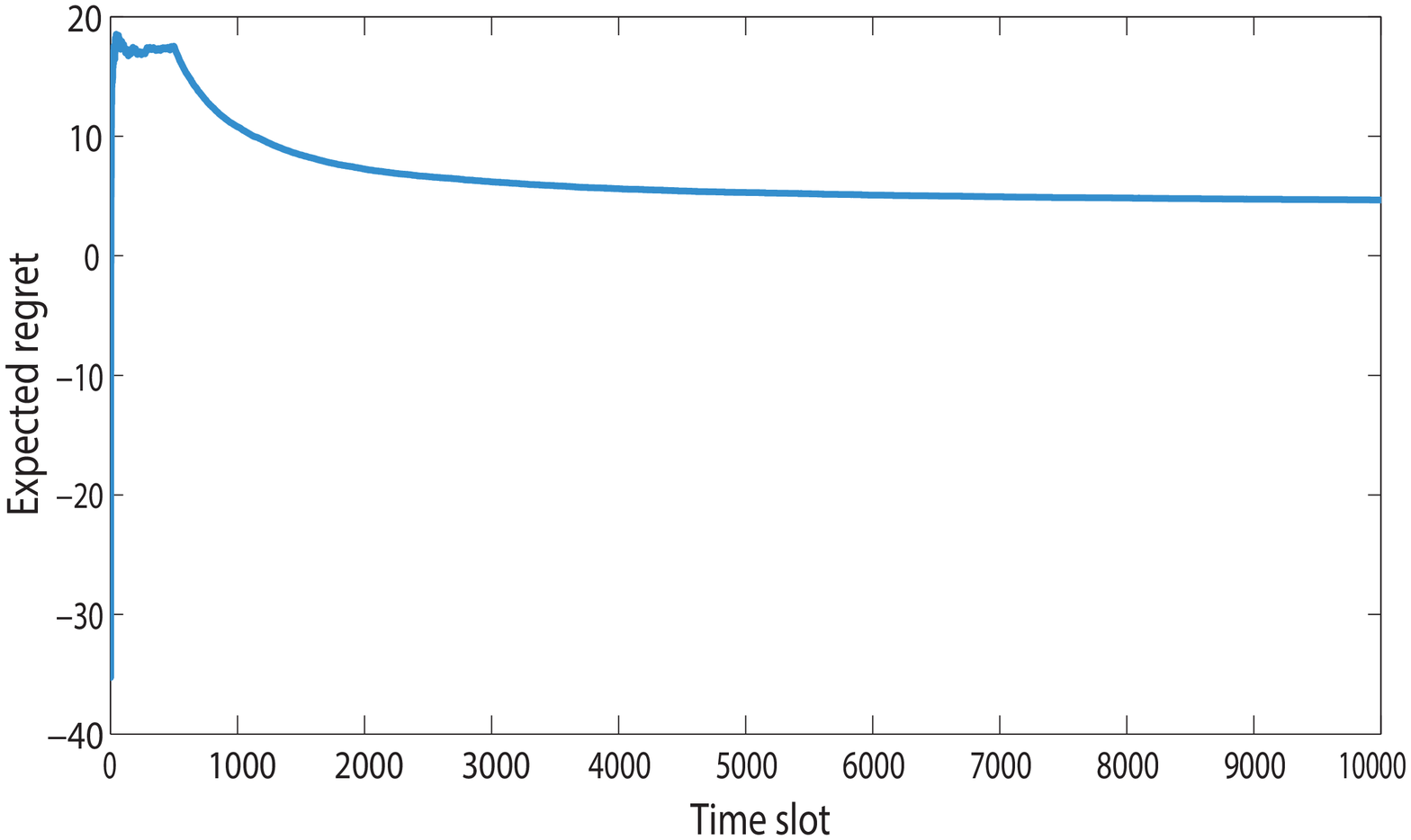}
                \label{fig-alg2-06}
                  \hfill
    }
  \caption{Expected regret of DFL-CSO}
  \label{fig-alg2}
\end{figure*}
Now we consider the combinatorial-play case with side reward.
Recall that in this scenario,
it requires to select a com-arm $\mathbf{s}_{x}$
with maximum side reward,
where  the side reward is
the sum of observed rewards
of all arms neighboring to arms in $\mathbf{s}_{x}$.
The case is more complicated than previous three cases,
due to:
1) Asymmetric observations on side reward for neighboring
nodes in one clique;
2) Probably exponential number of strategies caused
arbitrary constraint.
Therefore, it is complicated to analyze the
regret bound if adopting the same techniques of
combinatory-play with side observation.
Instead of learning side reward of strategies directly,
we learn the direct reward of arms that compose com-arms.

\begin{algorithm}[ht]
\caption{ Distribution-Free Learning policy for combinatorial-play with side reward (DFL-CSR)}
\label{alg-cwb}
  \begin{algorithmic}[1]
   \STATE For each time slot $t=0,1,\dots,n$ \\*
    Select a com-arm $\mathbf{s}_x$ by maximizing
        \begin{equation}
        \label{eqn-cwb}
         \sum_{i \in Y_x}
             \biggl( \overline{X}_{i,t}
                    +\sqrt{
                         \frac{ \max{(\ln{\frac{t^{2/3}}{K O_{i,t}} }},0)} {O_{i,t}}
                           }
             \biggl)
        \end{equation}
            to pull
    \STATE for $k \in Y_x$ do
        \STATE $O_{k,t+1}  \leftarrow  O_{k,t} +1 $
            \STATE $\overline{X}_{k,t+1} =
                \overline{X}_{k,t}/O_{k,t}^b+ (1-1/O_{k,t}^b)\overline{X}_{k,t}$
    \STATE end for
    \STATE end for
  \end{algorithmic}
\end{algorithm}

\begin{theorem}
\label{theorem-cwb}
    The expected regret of Algorithm~\ref{alg-cwb} after $n$ time slots is bounded by
\begin{eqnarray}
    \mathfrak{R}(n)
    &\leq& NK + \biggl(\sqrt{e K} + 8(1+N) N^3 \biggl) n^{\frac{2}{3}}
                \nonumber\\
   &&             + (1+ \frac{4 \sqrt{K}N^2}{e})  N^2 K n^{\frac{5}{6}}.
\end{eqnarray}
where $N \leq K$ is the maximum  of $|Y_x|, x=1\dots |F|$.
\end{theorem}

\begin{IEEEproof}
    See Appendix.
\end{IEEEproof}


\section{Simulation}
\label{sec-simu}
In this section, we evaluate the performance of
the proposed $4$ algorithms in simulations.
We mainly analyze the regret generated
by each algorithm after a long time slot $n=10000$.

We first evaluate regret generated by DFL-SSO,
and compare with MOSS learning policy.
The experiment setting is as follows.
We randomly generate a relation graph
   with $100$ arms, each following an i.i.d random process over time with mean between $[0,1]$.
We then plot the accumulated regret and expected regret over time, as shown in Fig.~\ref{fig-swb-exp}.
 Though the expected regret over time by MOSS converges to a value around $0$ that coincides with its theoretical bound in Fig.~\ref{fig-swb-exp}, it shows that its accumulated regret  grows dramatically.
It is oblivious the proposed algorithm with side information performs much better than MOSS, e.g., the accumulated regret and expected regret of our proposed algorithm (DFL-SSO) both converge to $0$.

\begin{figure}[!ptb]
    \centering
          \includegraphics[width=2.5in]{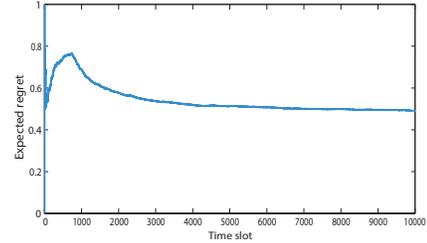}
                  \hfill
  \caption{Expected regret of DFL-SSR}
                \label{fig-alg3}
\end{figure}

\begin{figure}[!ptb]
    \centering
          \includegraphics[width=2.5in]{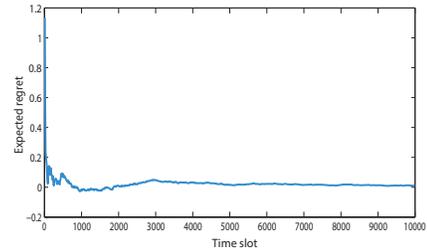}
                  \hfill
  \caption{Expected regret of DFL-CSR}
                \label{fig-alg4}
\end{figure}

For other $3$ algorithms, as we first study the $3$ variants of MAB problem, there are no candidate algorithms to compare.
We show the trend of expected regret over time for each case.
In evaluation of Algorithm~2, we note that the regret bound
contains the terms: number of com-arms and number of cliques.
The upper bound becomes huge if the number of com-arms is voluminous, and a small clique number can significantly
reduce the bound.
In order to investigate the impact experimentally, we then test for regret both under sparse relation graph and dense relation graph.
In Fig.~\ref{fig-alg2-03}, where the arms are uniformly and randomly connected with a
low probability of $0.3$, it shows that the expected regret slowly
increases beyond $0$.
While in Fig.~\ref{fig-alg2-06}, where the arms are uniformly and randomly connected with a higher probability of $0.6$,
it shows that the expected regret gradually approaches $0$.
It implicates that the side observation indeed helps
to reduce regret if one can observe more, even for the case
that previous literature show that it will introduce
exponential regret by learning each individual com-arm of a
huge feasible strategy set\cite{gai2012mab}.
The simulation results for Algorithm~3 and 4 are shown
in Fig.~\ref{fig-alg3} and \ref{fig-alg4},
where the expected regret in both figures
converges to $0$ dramatically.

\section{Related works}
\label{sec-review}
The classical multi-armed bandit problem does not assume
that existence of side bonus.   More recently, \cite{mannor2011bandits} and \cite{buccapatnam2014stochastic} considered the networked bandit problem in the presence of side observations. They study single play case and propose several policies whose regret bound depends on $\Delta_{\min}$, \eg, an arbitrarily small $\Delta_{\min}$ will invalidate the zero-regret result. In this work, we present the first distribution free policy for single play with side observation case.

For the variant with combinatorial play without side bonus, Anantharam et al. \cite{anantharam1987asymptotically} firstly consider the problem that exactly $N$ arms are selected simultaneously without constraint among arms.
Gai et al. recently extend this version to a more general problem with arbitrary constraints \cite{gai2012mab}.
The model is also relaxed to a linear combination of no more than $N$ arms.
However, the results presented in \cite{gai2012mab} are distribution-dependent.
To this end, we are the first to study combinatorial play case in the presence of side bonus. In particular, for the combinatorial play with side observation case, we develop a distribution-free zero regret learning policy. We theoretically show that this scheme converges faster than existing method. And for the combinatorial play with side reward case, we propose the first  distribution-free learning policy that has zero-regret.

\section{Conclusion}
\label{sec-con}
In this paper, we investigate networked combinatorial bandit problems under four cases.  This is motivated by the existence of potential correlation or influence among neighboring arms. We present and analyze a series of zero regret polices for each case. In the future, we are interested in investigating some heuristics
to improve the received regret in practice.
For example, at each time slot, instead of
playing the selected arm/strategy
 with maximum index value
 (Equation (\ref{eqn-swo}),
    (\ref{eqn-cwo})),
we will play the arm/strategy that
has maximum experimental average observation
among the neighbors of $I_t$.
Therefore, we ensure that
the received reward is better than the one with maximum index value.

\bibliographystyle{IEEEtran}
\bibliography{amab}

\section{Appendix}
\subsection{Proof of Theorem \ref{theorem-cwb}}

To prove the theorem, we will use Chernoff-Hoeffding bound and the maximal inequality by Hoeffding \cite{hoeffding1963probability}.

\begin{lemma} {(Chernoff-Hoeffding Bound \cite{hoeffding1963probability})}
\label{lemma-chernoff}
    $\xi_1,\dots,\xi_n$ are random variables within range $[0,1]$, and $E[\xi_t|\xi_1,...,\xi_{t-1}]=\mu,\forall 1 \leq t \leq n$. Let $S_n =\sum \xi_i$, then for all $a>0$
    {\small{
    \begin{eqnarray}
        \mathbf{P}(S_n \geq n \mu + a) \leq \exp{(-2 a^2 /n)},    \nonumber \\
        \mathbf{P}(S_n \leq n \mu - a) \leq  \exp{(-2 a^2 /n)}.
    \end{eqnarray}
    }}
\end{lemma}

\begin{lemma}{(Maximal inequality)}\cite{hoeffding1963probability}
\label{lemma-maximal}
    $\xi_1,\dots,\xi_n$ are i.i.d random variables with expect $\mu$, then for any $y>0$ and $n>0$,
    {\small{
    \begin{equation}
        \mathbf{P}  \biggl(
                    \exists \tau \in {1,\dots,n}, \sum_{t=1}^{\tau} (\mu-\xi_t) >y  \biggl)< \exp(-\frac{2y^2}{n})
            .
    \end{equation}
    }}
\end{lemma}

Each com-arm $\mathbf{s}_x$ and its neighboring arm set $Y_x$ actually compose a new com-arm,
which could be denoted by $Y_x$ as $\mathbf{s}_x \subset Y_x$.
Each new com-arm $Y_x$ corresponds to a unknown bonus $CB_{x,t}$ with mean $\sigma_{x}$.
Recall that we have assumed $\sigma_1 \geq \dots \geq \sigma_{|F|}$. As com-arm $Y_1$ is the optimal com-arm, we have $\Delta_x=\sigma_1-\sigma_x$, and let $Z_x = \sigma_1- \frac{\Delta_x}{2}$. We further define
$W_1 = \min_{1\leq t \leq n} W_{1,t}.$
We may assume the first time slot $z= \argmin_{1\leq t \leq n} W_{1,t}$.

\textbf{1. Rewrite regret in terms of arms}

Separating the strategies in two sets by $\Delta_{x_0}$ of some com-arm $\textbf{s}_{x_0}$(we will define $x_0$ later in the proof), we have
\setlength\arraycolsep{2pt}
\begin{eqnarray}
    \mathfrak{R}_{n} &=&    \sum_{x=1}^{x_0} \Delta_x E[T_{x,n}] + \sum_{x=x_0+1}^{|F|} \Delta_x E[T_{x,n}] \nonumber\\
   \label{rn}             &\leq&  \Delta_{x_0}n + \sum_{x=x_0+1}^{|F|} \Delta_x E[T_{x,n}] .
\end{eqnarray}

We then analyze the second term of (\ref{rn}). As there may be exponential number of strategies, counting $T_{x,n}$ of each com-arm by the classic upper-confidence-bound analysis yields regret growing linearly with the number of strategies. Note that each com-arm consists of $N$ arms at most, we can rewrite the regret in terms of arms instead of strategies. We then introduce a set of counters $\{\widetilde{T}_{x,n}|k=1,\dots,K\}$. At each time slot, either 1) a com-arm with $\Delta_x \leq \Delta_{x_0}$ or 2) a com-arm with $\Delta_x > \Delta_{x_0}$ is played. In the first case, no $\widetilde{T}_{x,n}$ will get updated. In the second case, we increase $\widetilde{T}_{x,n}$ by $1$ for any arm $k= \argmin_{j \in Y_x}\{ O_{j,t}\}$. Thus whenever a com-arm with $\Delta_x > \Delta_{x_0}$ is chosen, exactly one element in $\{\widetilde{T}_{x,n}\}$ increases by $1$. This implies that the total number that strategies of $\Delta_x > \Delta_{x_0}$ have been played is equal to sum of all counters in
$\{\widetilde{T}_{x,n}\}$, i.e., $\sum_{x=x_0+1}^{|F|} E[T_{x,n}] = \sum_{k=1}^{K} \widetilde{T}_{x,n}$. Thus, we can rewrite the second term of (\ref{rn}) as
{\small{
\setlength\arraycolsep{1pt}
\begin{eqnarray}
       \sum_{x=x_0+1}^{|F|} \Delta_x E[T_{x,n}]
   \leq \Delta_{X} \sum_{x=x_0+1}^{|F|}E[T_{x,n}]
\label{txtk}   &\leq& \Delta_{X} \sum_{k=1}^{K} E[ \widetilde{T}_{x,n}]. \nonumber\\
\end{eqnarray}
}}
Let ${I}_{k,t}$ be the indicator function that equals $1$ if $\widetilde{T}_{x,n}$ is updated at time slot $t$. Define the indicator function $\mathbf{1}\{y\} =1$ if the event $y$ happens and $0$ otherwise. When  ${I}_{k,t}=1$, a com-arm $Y_x$ with $x>x_0$ has been played for which $ O_{k,t} = \min \{O_{j,t}: \forall j \in Y_x \}$.
Then
{\small{
\setlength\arraycolsep{2pt}
\begin{eqnarray}
   \widetilde{T}_{x,n} &=& \sum_{t=1}^{n} \mathbf{1} \{{I}_{k,t}=1\} \\
                       &\leq& \sum_{t=1}^{n} \mathbf{1} \{W_{1,t} \leq W_{x,t}\} \\
                       &\leq& \sum_{t=1}^{n} \mathbf{1} \{W_1 \leq W_{x,t}\} \\
     \label{tk1}                  &\leq& \sum_{t=1}^{n} \mathbf{1} \{W_1 \leq W_{x,t}, W_1 \geq  Z_x \} \\
     \label{tk2}                  &~&       + \sum_{t=1}^{n} \mathbf{1} \{W_1 \leq W_{x,t}, W_1 < Z_x \} \\
     \label{tk}                 &=&  \widetilde{T}^{1}_{k,n} + \widetilde{T}^{2}_{k,n}.
\end{eqnarray}
}}
We use $\widetilde{T}^{1}_{k,n}$ and $\widetilde{T}^{2}_{k,n}$ to respectively denote Equation (\ref{tk1}) and (\ref{tk2}) for short. Next we show that both of the terms are bounded.

2. \textbf{Bounding $\widetilde{T}^{1}_{k,n}$}

Here we note the event  $\{W_1 \geq Z_x\}$ and $\{W_{x,t} > W_1 \}$ implies event $\{W_{x,t} > Z_x\}$. Let $\ln_+(y)= \max(\ln(y),0)$.
For any positive integer $l_0$, we then have,
\setlength\arraycolsep{2pt}
{\small{
\begin{eqnarray}
    \widetilde{T}^{1}_{k,n} &\leq& \sum_{t=1}^{n} \mathbf{1}\{W_{x,t} \geq Z_x\}  \\
                           &\leq& l_0 + \sum_{t=l_0}^{n} \mathbf{1}\{W_{x,t} \geq Z_x, \widetilde{T}^{1}_{k,t}> l_0\} \\
                           &=& l_0 + \sum_{t=l_0}^{n} \mathbf{P} \{W_{x,t} \geq Z_x, \widetilde{T}^{1}_{k,t}> l_0\} \\
                           &=& l_0 + \sum_{t=l_0}^{n}
     \label{tk1p3}                     \mathbf{P} \biggl\{ \sum_{j \in Y_x} \biggl(\overline{X}_{j,t} + \sqrt{\frac{\ln_+( \frac{t^{2/3}}{K O_{j,t}})} {l_0}} \biggl) \nonumber\\
                           &~&                \geq \sum_{j \in Y_x} \mu_j+ \frac{\Delta_x}{2}, \widetilde{T}^{1}_{k,t}> l_0
                                          \biggl\}.
\end{eqnarray}
}}
The event $\biggl\{ \sum_{j \in  Y_x} \biggl( \overline{X}_{j,t} + \sqrt{\frac{\ln_+(t^{2/3}/KO_{j,t})}{O_{j,t}}} \biggl) \geq \sum_{j \in Y_x} \mu_j+ \frac{\Delta_x}{2} \biggl\} $
indicates that the following must be true,
{\small{
\begin{equation}
    \exists j \in Y_x, \overline{X}_{j,t} + \sqrt{\frac{\ln_+(t^{2/3}/K O_{j,t})}{O_{j,t}}} \geq  \mu_j+ \frac{\Delta_x}{2N}.
\end{equation}
}}

Using union bound one directly obtains:
\setlength\arraycolsep{2pt}
\begin{eqnarray}
    \widetilde{T}^{1}_{k,n} &\leq& l_0 +  \sum_{t=l_0}^{n} \sum_{j \in Y_x}
                                \mathbf{P} \biggl\{ \overline{X}_{j,t} + \sqrt{\frac{\ln_+(t^{2/3}/K O_{j,t})}{O_{j,t}}}
                 \nonumber \\
          &&       \geq  \mu_j+ \frac{\Delta_x}{2N} \biggl\} \\
                        \label{tk1p1}     &\leq& l_0 +  \sum_{t=l_0}^{n} \sum_{j \in Y_x}   \mathbf{P} \biggl\{ \overline{X}_{j,t} - \mu_j
                   \nonumber \\
       &&                    \geq  \frac{\Delta_x}{2N}  - \sqrt{\frac{\ln_+(t^{2/3}/KO_{j,t})}{O_{j,t}}} \biggl\}.
\end{eqnarray}

Now we let $l_0=  16N^2 \lceil \ln(\frac{n^{3/4}}{K} \Delta_{x}^2) / \Delta_{x}^2) \rceil $ with $\lceil y \rceil$ the smallest integer larger than $y$. We further set $ \delta_0 = e^{1/2} \sqrt{K/n^{2/3}}$ and set $x_0$ such that $\Delta{x_0} \leq \delta_0 < \Delta_{x_0 +1}$.
As $O_{j,t} \geq l_0$,
{\small{
\begin{eqnarray}
    & &\ln_+ \biggl(\frac{t^{3/4}}{K O_{j,t}} \biggl)
                                \leq \ln_+ \biggl(\frac{n^{3/4}}{K O_{j,t}} \biggl)
                                \leq \ln_+ (n^{3/4}/K l_0)                              \nonumber \\
                                 & \leq & \ln_+ ( \frac{n^{3/4}}{K} \times \frac{\Delta_x^2}{16N^2} )
                                   \leq\frac{l_0 \Delta_x^2}{16N^2} \leq \frac{O_{j,t} \Delta_x^2}{16N^2}.
\end{eqnarray}
}}
Hence we have,
{\small{
\begin{equation}
    \frac{\Delta_x}{2N} - \sqrt{\frac{\ln_+(t^{3/4}/KO_{j,t})}{O_{j,t}}} \geq \frac{\Delta_x}{2N} - \frac{\Delta_x}{\sqrt{16N^2}} = c \Delta_x
\end{equation}
}}
with $c=\frac{1}{2N}-\frac{1}{\sqrt{16N^2}} = \frac{1}{4N}$.

Therefor, using Hoeffding's inequality and Equation (\ref{tk1p1}), and then plugging into the value of $l_0$, we get,

\begin{eqnarray}
\widetilde{T}^{1}_{k,n} &\leq & l_0 + \sum_{t=l_0}^{n} \sum_{j \in Y_x}   \mathbf{P}
                                \biggl\{ \overline{X}_{j,t} - \mu_j  \geq c \Delta_x \biggl\} \nonumber\\
                        &\leq& l_0 + \sum_{t=l_0}^{n} \sum_{j \in Y_x}  \exp(-2O_{j,t}(c\Delta_x)^2) \nonumber\\
                        &\leq& l_0 + K \cdot n \cdot  \exp(-2l_0(c\Delta_x)^2) \nonumber \\
   \label{tk1p2}        &=& 1+ 16N^2 \frac{\ln(\frac{n^{3/4}}{K} \Delta_{x}^2)}{ \Delta_{x}^2}+  K \cdot n \cdot \exp(-2\ln(n^{\frac{1}{12}}e)).
   \nonumber \\
\end{eqnarray}

 As $\delta_0 = e^{1/2} \sqrt{K/n^{\frac{2}{3}}}$ and $\Delta_x >\delta_0$, the second term in (\ref{tk1p2}) is bounded by \[\frac{16N^2(1+ \ln n^{1/12}) }{Ke} \cdot n^{2/3} < \frac{16N^2(n^{2/3}+ n^{3/4}) }{Ke} \]

The last term of (\ref{tk1p2}) is bounded by
\[K \cdot n \cdot \exp(-2\ln(n^{\frac{1}{12}}e)) \leq \frac{K}{e^2}\cdot n^{\frac{5}{6}}\]
Finally we get
{\small{
\begin{eqnarray}
    \widetilde{T}^{1}_{k,n} = 1+  \frac{16N^2(n^{2/3}+ n^{3/4}) }{Ke} +   \frac{K}{e^2}\cdot n^{\frac{5}{6}}.
\end{eqnarray}
}}

3. \textbf{Bounding $\widetilde{T}^{2}_{k,n}$}
{\small{
\begin{eqnarray}
    \widetilde{T}^{2}_{k,n}  &=&     \sum_{t=1}^{n} \mathbf{1}\{ W_1 \leq W_{x,t}, W_1 <  Z_x \}                      \nonumber \\
                            &\leq&  \sum_{t=1}^{n} \mathbf{P} \{ W_1 < Z_x\}
  \label{tk2n}                          \leq  n\mathbf{P} \{ W_1 < Z_x\}.
\end{eqnarray}
}}
Remember that at time slot $z$, we have $W_1 =\min{W_{1,t}}$. For  the probability $\{W_1 < Z_x\}$ of fixed $x$, we have
{\small{
\begin{eqnarray}
                    &~&     \mathbf{P}  \{W_1 < \sigma_1 - \frac{\Delta_x}{2}\}                                                                       \\
      &=&     \mathbf{P}  \biggl\{\sum_{j \in N_1, j=1}^{N} w_{j,z}<  \sigma_1 - \frac{\Delta_x}{2} \biggl\}           \\
 \label{Tk2p1} &\leq&  \sum_{j \in N_1} \mathbf{P} \biggl\{w_{j,z} <  \mu_{j } - \frac{\Delta_x}{2N}   \biggl\}.
    \end{eqnarray}
}}

We define function $f(u)= e \ln(\sqrt{\frac{n^{1/3}}{K}} u)/u^3$ for $u \in [\delta_0,N]$. Then we have,
{\small{
\begin{eqnarray}
   &~&  \mathbf{P}  \biggl\{w_{j,z} < \mu_{j } - \frac{\Delta_x}{2N}    \biggl\}          \nonumber \\                 & =&  \mathbf{P}  \biggl\{\exists 1 \leq l \leq n: \sum_{\tau=1}^{l} \biggl(X_{j,\tau}
              + \sqrt{\frac{\ln_+(\frac{\tau^{2/3}}{K l})} {l} } \biggl)                   < l\mu_{j } - \frac{l\Delta_x}{2N}  \biggl\}                                                    \nonumber \\
   &\leq& \mathbf{P}    \biggl\{
                       \exists 1 \leq l \leq n: \sum_{\tau=1}^{l} (\mu_{j }- X_{j,\tau})    >   \sqrt{l\ln_+(\frac{\tau^{2/3}}{K l})} +  \frac{l\Delta_x}{2N}
                \biggl\}                                                                                        \nonumber \\
   &\leq& \mathbf{P}    \biggl\{
                        \exists 1 \leq l \leq f(\Delta_x):                                                             \sum_{\tau=1}^{l} (\mu_{j }- X_{j,\tau})
                        >\sqrt{l\ln_+(\frac{\tau^{2/3}}{K l})}
                \biggl\}                                                                                         \nonumber \\
   &~& +  \mathbf{P}    \biggl\{
                        \exists f(\Delta_x) < l \leq n:                                                             
                            \sum_{\tau=1}^{l} (\mu_{j }- X_{j,\tau})
                        > \frac{l\Delta_x}{2N}
                \biggl\}.
\end{eqnarray}
}}
For the first term we use a peeling argument with a geometric grid of the form $\frac{1}{2^{g+1}} f(\Delta_{x}) \leq l \leq \frac{1}{2^{g}} f(\Delta_{x})$:
{\small{
\begin{eqnarray}
    &&    \mathbf{P}      \biggl\{
                        \exists 1 \leq l \leq f(\Delta_x): \sum_{\tau=1}^{l} (\mu_{j }- X_{j,\tau})       \nonumber\\
    &&               >\sqrt{l\ln_+(\frac{\tau^{2/3}}{K l})}
                \biggl\}                                                                                                   \nonumber\\
    &\leq& \sum_{g=0}^{\infty} \mathbf{P}\biggl\{
                       \exists  \frac{1}{2^{g+1}} f(\Delta_{x}) \leq l \leq \frac{1}{2^{g}} f(\Delta_{x}):               \sum_{\tau=1}^{l} (\mu_{j }- X_{j,\tau})
                    \nonumber \\
      &~&              >\sqrt{ \frac{f(\Delta_x)}{2^{g+1}}\ln_+(\frac{\tau^{2/3}2^g}{K f(\Delta_{x})})}
                \biggl\}                                                                                                       \nonumber\\
    &\leq&   \sum_{g=0}^{\infty} \exp   \biggl(
                                               -2\frac{ f(\Delta_x) \frac{1}{2^{g+1}} \ln_+(\frac{\tau^{2/3} 2^g}{K f(\Delta_x)})} {f(\Delta_x)\frac{1}{2^g}}
                                        \biggl)                                                                                 \nonumber\\
    &\leq&     \sum_{g=0}^{\infty} \biggl[
                                        \frac{K f(\Delta_x)}{n^{2/3}} \frac{1}{2^g}
                                \biggl]
    \leq
 \label{tk2r1}                                       \frac{2 K f(\Delta_x)}{n^{2/3}}
\end{eqnarray}
}}
where in the second inequality we use  Lemma~\ref{lemma-maximal}.

As the special design of function $f(u)$, we have $f(u)$ takes maximum of $\frac{n^{1/2}}{3 K^{3/2}} $ when $u=e^{1/3}\sqrt{K/n^{1/3}}$.
For $\Delta_x  >  e^{1/3}\sqrt{K/n^{1/3}}$ , we have
{\small{
\begin{eqnarray}
      \frac{2K f(\Delta_x)}{n^{2/3}} \leq \frac{2}{3 \sqrt{K}} n^{-1/6}.
\end{eqnarray}
}}
For the second term we also use a peeling argument but with a geometric grid of the form $2^g f(\Delta_x) \leq l < 2^{g+1} f(\Delta_x)$:
{\small{
\begin{eqnarray}
    &&  \mathbf{P}      \biggl\{
                        \exists f(\Delta_x) < l \leq n: \sum_{\tau=1}^{l} (\mu_{j }- X_{j,\tau})
                        > \frac{l\Delta_x}{2N}
                \biggl\}                                                                                                    \nonumber\\
    &\leq&      \sum_{g=0}^{\infty} \mathbf{P}\biggl\{
                                        \exists 2^{g} f(\Delta_x) \leq l \leq 2^{g+1} f(\Delta_x):                                                             \sum_{\tau=1}^{l} (\mu_{j }- X_{j,\tau})
                \nonumber \\
      &~&          > \frac{2^{g-1}f(\Delta_x)\Delta_x}{N}
                                    \biggl\}                                                                                \nonumber\\
    &\leq&      \sum_{g=0}^{\infty} \exp    \biggl(
                                                   \frac{ -2^g f(\Delta_x) \Delta_x^2}{ 4N^2}
                                            \biggl)                                                                         \nonumber\\
    &\leq&      \sum_{g=0}^{\infty} \exp    \biggl(
                                                    -(g+1)f(\Delta_x) \Delta_x^2 / 4N^2
                                            \biggl)                                                                         \nonumber\\
  \label{tk2r2}  &=&         \frac{1}{\exp(f(\Delta_x)\Delta_x^2/4N^2)-1}.
\end{eqnarray}
}}

We note that $f(u)u^2$ has a minimum of $\frac{e}{\sqrt{K}} n^{1/6}$ when $u=x_0$.
Thus for (\ref{tk2r2}), we further have,

\begin{eqnarray}
          \frac{1}{\exp(\frac{f(\Delta_x)\Delta_x^2}{4N^2})-1}  
    \leq  \frac{1}{\exp \bigg( \frac{ e n^{1/6}}{4 \sqrt{K}N^2}   \biggl) -1} 
    \leq  \frac{4 \sqrt{K}N^2 n^{-\frac{1}{6}}}{e}. \nonumber \\
\end{eqnarray}

Combining (\ref{Tk2p1}) and (\ref{tk2n}), we then have
{\small{
\begin{equation}
    \widetilde{T}^{2}_{k,n} \leq \frac{2N n^{5/6} }{3 \sqrt{K}} + \frac{4 \sqrt{K}N^3 n^{5/6}}{e}  \leq (1+\frac{4 \sqrt{K}N^2}{e})N n^{\frac{5}{6}}.
\end{equation}
}}

\textbf{4. Results without dependency on $\Delta_{\min}$}

Summing $\widetilde{T}^{1}_{k,n}$ and $\widetilde{T}^{2}_{k,n}$,
we have
\begin{eqnarray}
    \widetilde{T}_{x,n}  &\leq& \widetilde{T}^{1}_{k,n} + \widetilde{T}^{2}_{k,n}  \nonumber \\
                        &=&  1 + \frac{16N^2}{Ke} (1+ \frac{8N}{15}) n^{\frac{2}{3}} + (1+ \frac{4 \sqrt{K}N^2}{e}) N n^{\frac{5}{6}} \nonumber
\end{eqnarray}

and using  $\Delta_X \leq N$ and $\Delta_x \leq \delta_0$ for $x \leq x_0$, we have
\begin{eqnarray}
    \mathfrak{R}(n) &\leq& \sqrt{ K e} n^{\frac{2}{3}} + NK\biggl[1+ \frac{16N^2}{Ke} (1+ \frac{8N}{15}) n^{\frac{2}{3}}     \nonumber\\
    &~&
    +  (1+ \frac{4 \sqrt{K}N^2}{e} )N n^{\frac{5}{6}}  \biggl]                \nonumber\\
               &\leq& NK + \biggl(\sqrt{e K} + 8(1+N) N^3 \biggl) n^{\frac{2}{3}}
               \nonumber\\
      &&          + (1+ \frac{4 \sqrt{K}N^2}{e})  N^2 K n^{\frac{5}{6}}.
    \nonumber
\end{eqnarray}

\end{document}